\documentclass[conference]{IEEEtran}
\IEEEoverridecommandlockouts
\usepackage{cite}
\usepackage{amsmath,amssymb,amsfonts}
\usepackage{graphicx}
\usepackage{textcomp}
\usepackage{xcolor}
\usepackage{booktabs}
\usepackage{multirow}

\begin{document}

\title{OAA: Three Phases of Vocal Guidance in Human-Drone Teleoperation}

\author{
\IEEEauthorblockN{
Allan Henry\textsuperscript{1,2,3},
Christian Graff\textsuperscript{3},
Solange Rossato\textsuperscript{2},
Jos\'e-Ernesto Gomez-Balderas\textsuperscript{1},
Sylvain Huet\textsuperscript{1}}
\IEEEauthorblockA{
\textsuperscript{1}GIPSA-Lab,
\textsuperscript{2}LIG,
\textsuperscript{3}LPNC,
Univ. Grenoble Alpes, Grenoble, France\\
firstname.lastname@univ-grenoble-alpes.fr}
}

\maketitle

\begin{abstract}
Voice-guided teleoperation requires systems that adapt to the evolving dynamics of human guidance. Yet most voice-controlled robot systems treat spoken commands as a stationary stream, ignoring how the guide's communicative behavior changes as the task progresses. Using motion capture and speech data from two experimental configurations, human-human guidance (finger pointing, $N{=}10$ dyads) and human-drone teleoperation (gamepad control, $N{=}29$ dyads), we show that spontaneous vocal guidance consistently organizes into three kinematically and linguistically distinct phases: Orientation, Approach, and Adjustment. These phases are identified automatically via change point detection on 3D trajectory signals, and validated statistically (Kruskal-Wallis, $p < .001$). Three lexical families replicate across configurations: rotation vocabulary marks Orientation, translation vocabulary is scarce there, and attenuators accumulate toward Adjustment. Together with inter-utterance silence, these cues mark the Orientation boundary that speech rate alone leaves unmarked. The same three-phase structure emerges in both configurations despite radically different motor interfaces, suggesting it is an intrinsic property of human spatial guidance rather than an artifact of the experimental setup. We discuss implications for OAA-aware adaptive control in voice-guided teleoperation.
\end{abstract}


\section{Introduction}

Voice control is a promising modality for drone teleoperation, especially for non-expert operators who cannot master traditional gamepad interfaces~\cite{tezza2019, hunt2025, suarez2016}. However, most existing voice-controlled UAV systems rely on predefined command vocabularies and treat each spoken command independently, applying fixed movement durations and context-free classification~\cite{fayjie2017, choutri2022, contreras2020}. This design implicitly assumes that the guide's communicative behavior is stationary throughout the task, an assumption that does not hold in practice.

In naturalistic settings, human guidance is a dynamic process. The way a guide formulates instructions, the frequency at which they speak, and the duration of the silences between commands all evolve as the guided entity progresses toward its goal. Prior work on task-oriented dialogue has shown that speech production in situated interaction is tightly coupled with the dynamics of the physical task~\cite{stoia2008, thompson1993}. In the specific context of voice-guided drone navigation, a preliminary qualitative analysis~\cite{henry2025petra} suggested that guidance unfolds in three recurring phases (orientation, approach, and adjustment) each characterized by distinct movement patterns and lexical choices. However, these phases were identified manually and lacked statistical validation.

This observation motivates three questions that the present paper addresses. First, can these phases be detected algorithmically from trajectory data alone, without manual annotation? Second, does the same three-phase structure hold when the guided entity is a physical drone rather than a human? Third, does the guide's language change qualitatively across phases, or only in its temporal distribution? To answer these questions, we leverage data from the VoiceStick corpus~\cite{henry2026taln}, a French spontaneous speech dataset collected during real human-drone teleoperation sessions, alongside an earlier human-human guidance dataset collected under the same spatial layout. We apply change point detection to 3D motion capture data from both configurations, validate the resulting phases against a baseline segmentation, and analyze the temporal and lexical properties of each phase.

\section{Materials and Methods}

\subsection{General Protocol}

Both configurations share the same asymmetric guide-pilot paradigm. A \textit{guide}, located in a separate room, provides spoken instructions to direct a \textit{pilot} toward invisible 3D targets. The targets are positioned within a $3 \times 3 \times 3$ grid of 27 possible locations (spacing: 135~cm horizontally, heights: 70, 110, and 150~cm). The guide observes a virtual reconstruction of the environment displayed via Unreal Engine, showing the target as a sphere with a 15~cm validation radius. Communication is strictly unidirectional: only the guide speaks. The pilot receives no visual information about the target and relies exclusively on the guide's instructions.

Positions and orientations were recorded at 100~Hz by a 12-camera VICON motion capture system. The guide's speech was recorded continuously at 16~kHz via a lavalier microphone.

The two configurations were chosen to dissociate the guide's communicative strategy from the pilot's execution channel. Everything on the guide's side is held constant: spatial layout, virtual reconstruction, unidirectional protocol, and the task of directing a pilot who cannot see the target. Everything on the pilot's side differs maximally, from embodied self-motion with intact proprioception and no latency to remote actuation subject to inertia, latency and a narrow camera view. An OAA structure arising from drone dynamics should be absent from Cfg.~A; one arising from walking and pointing biomechanics should be absent from Cfg.~B. Its presence in both localizes it in what remains invariant, namely the guide's planning. The contrast opposes two packages of conditions rather than manipulating one factor, and therefore establishes invariance without isolating causes.

\subsection{Configuration A: Human-Human Guidance}

In Configuration~A, the pilot is a human participant who stands in the capture volume with eyes closed, holding an optical marker at the tip of their index finger. The pilot moves by walking and pointing, using their body as the navigating entity. Ten dyads of French-speaking participants (undergraduate psychology students with no prior experience in guidance tasks) each completed multiple target reaches, producing a total of 120 trials.

\subsection{Configuration B: Human-Drone Teleoperation}

In Configuration~B, the pilot controls a DJI Tello micro-drone using an Xbox gamepad. The pilot observes the environment through the drone's first-person view (FPV) camera feed displayed on a monitor and has no direct line of sight to the drone or the targets. Twenty-nine dyads of French-speaking participants completed target reaches under the same spatial layout, producing 338 analyzed trials. The speech and gamepad data from this configuration form the VoiceStick corpus~\cite{henry2026taln}, a publicly available resource for research on voice-guided drone interaction.

\subsection{Phase Segmentation}

For each trial, three time-varying signals were computed from the VICON position data: Euclidean distance between the guided entity (fingertip or drone) and the target (cm), radial approach speed (first derivative of distance, cm/s), and yaw angular velocity (rad/s). Signals were smoothed with a 500~ms moving average (50 samples) to attenuate measurement noise, then Z-normalized so that linear and angular variations contribute equally to the analysis.

A Binary Segmentation algorithm~\cite{truong2020} with $\ell_2$ cost and a minimum segment size of 50 samples (0.5~s) was applied to the resulting 3D signal. The algorithm was configured to detect exactly two change points per trial, producing three chronological segments that we refer to as the OAA phases: Orientation~(O), Approach~(Ap), and Adjustment~(Aj). The choice to fix the number of change points to two follows from the qualitative observations of~\cite{henry2025petra} and the principle of parsimony: a three-phase model captures the main regime changes without over-segmenting short trials.Because segmentation and phase comparison rely on the same signals, we also computed a control segmentation splitting each trial into three equal-duration thirds, and compared effect sizes ($\eta^2$ derived from the Kruskal-Wallis statistic) between the two partitions.

\subsection{Metrics and Statistical Analysis}

Three kinematic metrics were extracted per phase and per trial: mean radial velocity (cm/s), mean yaw angular velocity (rad/s), and jerk (derivative of acceleration, cm/s$^3$). Phase duration was also recorded. Speech was segmented using PyAnnote~\cite{bredin2023} and transcribed with Whisper large-v2~\cite{radford2023}. Each utterance was assigned to a phase based on the timestamp of its temporal midpoint. Linguistic metrics include utterance count, verbal density (words per second, computed as total words over total phase duration) and mean inter-utterance silence duration (s). Transcriptions were further annotated for six semantic categories of guidance vocabulary, defined a priori from~\cite{henry2025petra} and matched by regular expressions over accent-normalized tokens: rotation (\textit{tourne} `turn', \textit{pivote} `pivot'), horizontal translation (\textit{avance} `move forward', \textit{recule} `move back', \textit{tout droit} `straight ahead'), vertical translation (\textit{monte} `go up', \textit{descends} `go down'), attenuation (\textit{un peu} `a little', \textit{l\'eg\`erement} `slightly', \textit{doucement} `gently'), stop and validation (\textit{stop}, \textit{voil\`a} `that's it', \textit{c'est bon} `good'), and deixis (\textit{l\`a} `there', \textit{devant toi} `in front of you'). No category relies on disfluencies, which Whisper normalizes. Category frequencies were expressed per 100 words within each phase of each trial, so that trials rather than word tokens form the unit of analysis, and compared across phases with Kruskal-Wallis tests and Holm correction over the six categories. Kinematic comparisons used Kruskal-Wallis tests (non-parametric, suited to the non-normal distributions of kinematic data), followed by Dunn post-hoc tests with Holm correction ($\alpha = .05$).

\section{Results}

\subsection{Kinematic Validation}

Table~\ref{tab:kinematic} reports the kinematic results for both configurations, confirming that the OAA phases correspond to distinct movement regimes.

\begin{table}[ht]
\centering
\caption{Kinematic characteristics per phase (M $\pm$ SD). Superscript letters indicate Dunn post-hoc groupings ($p < .05$, Holm correction). Phases sharing a letter do not differ significantly.}
\label{tab:kinematic}
\setlength{\tabcolsep}{3pt}
\footnotesize
\begin{tabular}{llccc}
\toprule
& & \textbf{O} & \textbf{Ap} & \textbf{Aj} \\
\midrule
\multirow{4}{*}{\rotatebox{90}{\scriptsize Cfg. A}}
& Vel. (cm/s)
  & $10.00 \pm 9.66^a$
  & $17.96 \pm 12.35^b$
  & $5.61 \pm 4.57^c$ \\
& Ang. vel. (rad/s)
  & $0.66 \pm 0.71^a$
  & $0.44 \pm 0.26^b$
  & $0.22 \pm 0.09^c$ \\
& Jerk (cm/s$^3$)
  & $20.5 \pm 34.6^a$
  & $19.0 \pm 33.5^a$
  & $2.7 \pm 3.9^b$ \\
& Duration (s)
  & $7.4 \pm 5.7$
  & $8.3 \pm 6.7$
  & $19.7 \pm 12.1$ \\
\midrule
\multirow{4}{*}{\rotatebox{90}{\scriptsize Cfg. B}}
& Vel. (cm/s)
  & $2.84 \pm 4.84^a$
  & $33.45 \pm 22.81^b$
  & $5.76 \pm 5.63^c$ \\
& Ang. vel. (rad/s)
  & $0.51 \pm 0.25^a$
  & $0.45 \pm 0.30^b$
  & $0.38 \pm 0.19^c$ \\
& Jerk (cm/s$^3$)
  & $10.1 \pm 71.9^a$
  & $35.8 \pm 119.4^b$
  & $10.6 \pm 36.7^c$ \\
& Duration (s)
  & $15.1 \pm 11.0$
  & $9.1 \pm 12.8$
  & $25.2 \pm 21.3$ \\
\bottomrule
\end{tabular}
\end{table}


Mean radial velocity distinguishes all three phases in both configurations ($p < .003$ for every pairwise comparison, and $p < 10^{-6}$ for five of the six). Velocity is low during Orientation, peaks during Approach as the workspace is traversed, and drops during Adjustment. The contrast is far sharper in Cfg.~B (2.8 to 33.5 cm/s) than in Cfg.~A (10.0 to 18.0), reflecting the drone's higher traversal speed and its near-immobility while the pilot establishes a heading.

Angular velocity decreases across the three phases in both configurations, every pairwise contrast being significant (all $p < .006$). Heading corrections thus become progressively smaller as the entity converges. The decline is gradual rather than stepwise: the control analysis below shows that an arbitrary partition separates angular velocity at least as well as the detected boundaries, so this variable tracks elapsed time rather than the phase structure. Rotation is therefore not confined to an initial orienting stage, consistent with coupled piloting, but it does not index the OAA phases.

Jerk is the one metric on which the configurations diverge. In Cfg.~A, Orientation and Approach are indistinguishable ($p = .68$) and only Adjustment stands apart, reflecting the smaller amplitudes of fine positioning. In Cfg.~B all three differ ($p < .001$), Approach carrying by far the highest jerk, as the drone's inertia, latency and gamepad discretization amplify acceleration variations during fast traversal.

These effects are not a by-product of the segmentation. Under the equal-thirds baseline, the effect of phase on radial velocity falls from $\eta^2 = .21$ to $.08$ in Cfg.~A and from $.50$ to $.08$ in Cfg.~B, and on jerk from $.28$ to $.02$ and $.20$ to $.01$. Angular velocity behaves differently: the baseline explains it at least as well as the detected partition ($.46$ vs $.31$; $.10$ vs $.09$). The detected boundaries thus capture genuine regime changes in translation and smoothness, but not in heading.

\subsection{Linguistic and Lexical Synchronization}

\begin{table}[ht]
\centering
\caption{Linguistic characteristics per phase. Lower block: mean per-trial frequency of each lexical category, in occurrences per 100 words. Asterisks mark categories differing significantly across phases (Kruskal-Wallis, Holm correction over six categories, $p < .05$).}
\label{tab:linguistic}
\setlength{\tabcolsep}{3pt}
\footnotesize
\begin{tabular}{llccc}
\toprule
& & \textbf{O} & \textbf{Ap} & \textbf{Aj} \\
\midrule
\multirow{9}{*}{\rotatebox{90}{\scriptsize Cfg. A}}
& Utterances & 180 & 236 & 759 \\
& Density (w/s) & 1.04 & 1.25 & 1.77 \\
& Silence (s) & 2.11 & 1.66 & 1.69 \\
\cmidrule(l){2-5}
& Rotation$^{*}$ & 2.95 & 0.77 & 1.01 \\
& Transl. horiz.$^{*}$ & 3.28 & 6.49 & 7.34 \\
& Transl. vert.$^{*}$ & 1.68 & 3.91 & 3.58 \\
& Attenuation$^{*}$ & 5.54 & 8.33 & 10.23 \\
& Stop / valid.$^{*}$ & 8.05 & 7.60 & 10.53 \\
& Deixis$^{*}$ & 2.75 & 3.63 & 4.03 \\
\midrule
\multirow{9}{*}{\rotatebox{90}{\scriptsize Cfg. B}}
& Utterances & 929 & 418 & 1866 \\
& Density (w/s) & 1.05 & 0.78 & 1.28 \\
& Silence (s) & 2.73 & 1.93 & 2.03 \\
\cmidrule(l){2-5}
& Rotation$^{*}$ & 3.81 & 2.30 & 1.19 \\
& Transl. horiz.$^{*}$ & 6.03 & 8.89 & 8.86 \\
& Transl. vert.$^{*}$ & 2.93 & 2.73 & 3.11 \\
& Attenuation$^{*}$ & 3.67 & 6.05 & 7.87 \\
& Stop / valid. & 11.21 & 12.17 & 12.68 \\
& Deixis & 3.65 & 3.04 & 2.43 \\
\bottomrule
\end{tabular}
\end{table}

Adjustment concentrates the bulk of the guide's speech in both configurations, 65\% of utterances in Cfg.~A and 58\% in Cfg.~B, and reaches the highest verbal density (1.77 and 1.28 words/s). Since it also occupies about half of each trial (56\% and 51\%), this reflects a genuine rise in speaking rate: utterance rate climbs from 13.5 to 21.4 per minute in Cfg.~A and from 10.9 to 13.1 in Cfg.~B.

The configurations differ in mid-trial. Cfg.~A shows a monotonic gradient, whereas in Cfg.~B Approach is the quietest phase of all (0.78 words/s), below Orientation. Orientation lasts twice as long in Cfg.~B as in Cfg.~A (15.1 against 7.4~s), the drone having to stabilize in hover before any useful displacement, so instructions accumulate; during Approach the drone then covers the workspace quickly and unaided while the guide falls silent. This is what inter-utterance silence as a predictor of movement duration predicts~\cite{henry2026icmi}: long displacements elicit long silences, short micro-corrections a rapid vocal response.

A natural question follows: can the OAA phases be detected from vocal features alone, without trajectory data? To test this, we applied the same segmentation algorithm to a signal constructed from speech density, silence duration, and utterance duration. The resulting breakpoints diverged substantially from the kinematic ones in both configurations (median gap 4.2~s and 7.9~s in Cfg.~A, 6.0~s and 9.9~s in Cfg.~B, i.e.\ 14--27\% of trial duration).

Taken individually, the vocal features mark different boundaries. Verbal density and utterance rate separate Adjustment sharply from the two earlier phases in both configurations ($p < .003$) while leaving Orientation and Approach indistinguishable ($p = .11$ and $p = .83$ for density). Inter-utterance silence shows the opposite profile, longest during Orientation and stable afterwards (2.73, 1.93 and 2.03~s in Cfg.~B, $p < .001$ against either later phase, $p = .33$ between them); the same ordering appears in Cfg.~A but falls just short of significance ($p = .053$) with 120 trials. Utterance duration and words per utterance vary weakly and inconsistently. Speech thus carries both boundaries, through two channels that a single joint segmentation fails to recover.

The lexicon marks a different boundary. Three categories replicate across configurations. Rotation vocabulary is densest during Orientation and falls afterwards (2.95 to 1.01 per 100 words in Cfg.~A, 3.81 to 1.19 in Cfg.~B). Horizontal translation shows the mirror-image profile, scarce during Orientation and roughly twice as frequent later (3.28 to 7.34 and 6.03 to 8.86). Attenuators accumulate steadily toward Adjustment and carry the largest effect of all six categories in both configurations (5.54 to 10.23, $\eta^2 = .09$; 3.67 to 7.87, $\eta^2 = .13$). The first two therefore mark the Orientation boundary that speech rate leaves unmarked, converging with the silence pattern reported above, while the third reinforces the Adjustment boundary. Two independent channels, one prosodic-temporal and one lexical, thus flag the same transition that rate features miss.

The other categories replicate less well: vertical translation is ordered only in Cfg.~A, stop markers reach significance only there, and deixis rises in Cfg.~A while falling and failing correction in Cfg.~B. Effect sizes are modest throughout ($\eta^2$ between .03 and .13), so the lexical signature is real and replicable but partial.

\section{Discussion}

\subsection{Generalization Across Configurations, Tasks, and Platforms}
\label{sec:generalization}

The convergence of results across two radically different motor interfaces (direct fingertip guidance vs.\ gamepad-controlled drone) provides evidence that the three-phase structure is an intrinsic property of human spatial guidance in 3D navigation, not an artifact of the experimental setup. Because the two configurations share only the guide's side of the interaction, this convergence localizes the phase structure in communicative planning rather than in the execution channel.

Interestingly, the Approach-to-Adjustment transition occurs at comparable distances in both configurations (96~cm in Cfg.~A, 109~cm in Cfg.~B). Rather than a purely motor threshold, this distance aligns with established communicational boundaries in spatial dialogue. Prior work has shown that as the distance to a target decreases toward arm's length (around 100~cm), human guides naturally alter their communicative strategy to achieve joint attention, suppressing broad indications in favor of denser, localized verbal descriptions~\cite{bangerterUsingPointingDescribing2004}. This mirrors the rise in verbal density we observe during Adjustment, and an analogous shift occurs in Cfg.~B although the pilot is physically remote, which suggests that the transition to micro-corrections is triggered by a perceptual and communicative boundary rather than by the pilot's reaching space. This parallel warrants further investigation with manipulations of target size and camera zoom to disentangle perceptual and motor contributions to the phase transition.

Generalization beyond this task can be reasoned about rather than merely speculated upon, because two of the three phases have a theoretical counterpart. The classical two-component model of aimed movement decomposes a goal-directed action into a ballistic impulse followed by feedback-driven correction near the target~\cite{woodworthAccuracyVoluntaryMovement1899, meyerOptimalityHumanMotor1988}, matching our Approach and Adjustment, with Orientation reflecting the additional need to establish a heading in an extended 3D workspace. OAA would then be the vocal signature of a generic organization of goal-directed action rather than a peculiarity of droning, which yields testable predictions. Multi-waypoint missions should produce one OAA cycle per waypoint, Orientation compressing whenever the next waypoint is already in view and its rotation vocabulary reappearing at each cycle. Obstacle avoidance adds intermediate subgoals and should fragment Approach into alternating traversal and correction episodes, favoring more than two change points under a penalized criterion. A manipulator reaching for an object should show Approach and Adjustment without a genuine Orientation phase, and without the rotation peak. Targets unknown in advance would prepend an exploration phase dominated by uncertainty markers rather than directives.

\subsection{Implications for Adaptive Teleoperation}

The identification of the OAA phases opens the door to phase-aware adaptive systems. A voice-controlled drone that estimates the current phase from trajectory features could adapt its behavior: accepting broader directional tolerance during Orientation, sustaining smooth continuous motion during Approach, and switching to high-frequency micro-correction mode during Adjustment. Estimation need not rely on the trajectory at all: long silences and a rotation-heavy, translation-poor lexicon signal that the guide is still orienting, while rising density and attenuator frequency signal entry into Adjustment. A lightweight classifier could track both transitions without telemetry, provided it treats them as separate detectors rather than one change point.

\subsection{Limitations and Future Work}
\label{sec:limitations}

The algorithm assumes exactly three phases, whereas obstacle avoidance or multi-waypoint navigation might exhibit more. Angular velocity, moreover, tracks elapsed time rather than the detected boundaries, so a two-channel variant restricted to distance and radial velocity would clarify whether it belongs in the segmentation at all.

The corpus is exclusively French-speaking, and cross-linguistic replication would strengthen the generalizability claim, particularly for lexical categories defined a priori over French vocabulary, which could also be induced from the data. Trials are nested within dyads and were treated as independent.

Manipulating target distance, drone dynamics, camera field of view and starting orientation would reveal which factors modulate the phase structure, turning the package comparison into a factorial design. Richer speech representations, such as learned SSL embeddings, might also capture phase-related variation that hand-crafted features miss.

\section{Conclusion}

We showed that spontaneous vocal guidance in 3D navigation consistently organizes into three phases (Orientation, Approach, and Adjustment) that differ in their kinematic and linguistic properties. This structure is robust across motor interfaces and emerges from data-driven segmentation without manual annotation. Speech marks the two transitions through different channels, verbal density isolating Adjustment while inter-utterance silence and rotation vocabulary isolate Orientation, which together make phase estimation from speech alone a realistic prospect.

The Adjustment phase concentrates the majority of communicative effort, making it the most critical phase for voice-controlled teleoperation systems. These findings motivate the development of OAA-aware adaptive control strategies for human-drone interaction.

\section*{Acknowledgment}
This work is supported by the French National Research Agency in the framework of the "Investissements d'avenir" program (ANR-15-IDEX-02) and has been partially supported by ROBOTEX 2.0 (Grants ROBOTEX ANR-10-EQPX-44-01 and TIRREX ANR-21-ESRE-0015) funded by the French program Investissements d'avenir. This work has benefited from collaboration with members of SAMGuide (ANR-21-VE33-0011-01). The authors would like to thank all participants who took part in the data collection experiment.

\clearpage
\bibliographystyle{IEEEtran}
\bibliography{references}

\begin{thebibliography}{10}
\providecommand{\url}[1]{#1}
\csname url@samestyle\endcsname
\providecommand{\newblock}{\relax}
\providecommand{\bibinfo}[2]{#2}
\providecommand{\BIBentrySTDinterwordspacing}{\spaceskip=0pt\relax}
\providecommand{\BIBentryALTinterwordstretchfactor}{4}
\providecommand{\BIBentryALTinterwordspacing}{\spaceskip=\fontdimen2\font plus
\BIBentryALTinterwordstretchfactor\fontdimen3\font minus \fontdimen4\font\relax}
\providecommand{\BIBforeignlanguage}[2]{{%
\expandafter\ifx\csname l@#1\endcsname\relax
\typeout{** WARNING: IEEEtran.bst: No hyphenation pattern has been}%
\typeout{** loaded for the language `#1'. Using the pattern for}%
\typeout{** the default language instead.}%
\else
\language=\csname l@#1\endcsname
\fi
#2}}
\providecommand{\BIBdecl}{\relax}
\BIBdecl

\bibitem{tezza2019}
D.~Tezza and M.~Andujar, ``The state-of-the-art of human--drone interaction: A survey,'' \emph{IEEE Access}, vol.~7, pp. 167\,438--167\,454, 2019.

\bibitem{hunt2025}
W.~Hunt, S.~D. Ramchurn, and M.~D. Soorati, ``A survey of language-based communication in robotics,'' \emph{arXiv preprint arXiv:2406.04086}, 2025.

\bibitem{suarez2016}
R.~A. Suarez~Fernandez, J.~L. Sanchez-Lopez, C.~Sampedro, H.~Bavle, M.~Molina, and P.~Campoy, ``Natural user interfaces for human-drone multi-modal interaction,'' in \emph{Proc. International Conference on Unmanned Aircraft Systems (ICUAS)}, 2016, pp. 1013--1022.

\bibitem{fayjie2017}
A.~R. Fayjie, A.~Ramezani, D.~Oualid, and D.~J. Lee, ``Voice enabled smart drone control,'' in \emph{Proc. Ninth International Conference on Ubiquitous and Future Networks (ICUFN)}, 2017, pp. 119--121.

\bibitem{choutri2022}
K.~Choutri, M.~Lagha, S.~Meshoul, M.~Batouche, Y.~Kacel, and N.~Mebarkia, ``A multi-lingual speech recognition-based framework to human-drone interaction,'' \emph{Electronics}, vol.~11, no.~12, p. 1829, 2022.

\bibitem{contreras2020}
R.~Contreras, A.~Ayala, and F.~Cruz, ``Unmanned aerial vehicle control through domain-based automatic speech recognition,'' \emph{Computers}, vol.~9, no.~3, p.~75, 2020.

\bibitem{stoia2008}
L.~Stoia, D.~M. Shockley, D.~K. Byron, and E.~Fosler-Lussier, ``Scare: A situated corpus with annotated referring expressions,'' in \emph{Proc. LREC}, 2008.

\bibitem{thompson1993}
H.~S. Thompson, A.~H. Anderson, E.~G. Bard, G.~Doherty-Sneddon, A.~Newlands, and C.~Sotillo, ``The {HCRC} map task corpus: Natural dialogue for speech recognition,'' in \emph{Proc. Workshop on Human Language Technology}, 1993, pp. 25--30.

\bibitem{henry2025petra}
A.~Henry, C.~Graff, S.~Rossato, J.-E. Gomez-Balderas, and S.~Huet, ``Voice commands for guidance to a {3D} position: To collect spontaneous data,'' in \emph{Proc. 18th ACM International Conference on PErvasive Technologies Related to Assistive Environments (PETRA)}, 2025, pp. 410--411.

\bibitem{henry2026taln}
\BIBentryALTinterwordspacing
A.~Henry, S.~Rossato, C.~Graff, S.~Huet, and J.-E. Gomez-Balderas, ``Voicestick : un corpus de parole spontanée pour le guidage vocal de drones,'' in \emph{Actes de CORIA-TALN 2026. Actes des 33ème Conférence sur le Traitement Automatique des Langues Naturelles. Volume 1 : articles scientifiques originaux}.\hskip 1em plus 0.5em minus 0.4em\relax Nantes, France: Association pour le Traitement Automatique des Langues, 6 2026, pp. 702--714. [Online]. Available: \url{https://talnarchives.atala.org/TALN/TALN-2026/41.pdf}
\BIBentrySTDinterwordspacing

\bibitem{truong2020}
C.~Truong, L.~Oudre, and N.~Vayatis, ``Selective review of offline change point detection methods,'' \emph{Signal Processing}, vol. 167, p. 107299, 2020.

\bibitem{bredin2023}
H.~Bredin, ``pyannote.audio 2.1 speaker diarization pipeline: Principle, benchmark, and recipe,'' in \emph{Proc. INTERSPEECH}, 2023, pp. 1983--1987.

\bibitem{radford2023}
A.~Radford, J.~W. Kim, T.~Xu, G.~Brockman, C.~McLeavey, and I.~Sutskever, ``Robust speech recognition via large-scale weak supervision,'' in \emph{Proc. ICML}, 2023, pp. 28\,492--28\,518.

\bibitem{henry2026icmi}
A.~Henry, S.~Rossato, C.~Graff, J.-E. Gomez-Balderas, and S.~Huet, ``Beyond rhythm and prosody: Inter-utterance silence predicts drone movement duration in spontaneous vocal guidance,'' in \emph{Proc. 28th ACM International Conference on Multimodal Interaction (ICMI)}, 2026, under review.

\bibitem{bangerterUsingPointingDescribing2004}
A.~Bangerter, ``Using pointing and describing to achieve joint focus of attention in dialogue,'' \emph{Psychological science}, vol.~15, pp. 415--9, 2004.

\bibitem{woodworthAccuracyVoluntaryMovement1899}
R.~S. Woodworth, ``Accuracy of voluntary movement,'' \emph{The Psychological Review: Monograph Supplements}, vol.~3, no.~3, pp. i--114, 1899.

\bibitem{meyerOptimalityHumanMotor1988}
D.~E. Meyer, R.~A. Abrams, S.~Kornblum, C.~E. Wright, and J.~E. Keith~Smith, ``Optimality in human motor performance: {{Ideal}} control of rapid aimed movements,'' \emph{Psychological Review}, vol.~95, no.~3, pp. 340--370, 1988.

\end{thebibliography}

\end{document}